\documentclass{article}
\usepackage{hyperref}

\usepackage{latexsym}
\usepackage{graphicx,grffile} 
\usepackage{mathrsfs} 
\usepackage{amsmath,amssymb,amsthm}  
\usepackage[american]{babel}
\usepackage{subcaption}
\usepackage{xcolor}
\usepackage[T1]{fontenc}
\usepackage[utf8]{inputenc}
\usepackage{algorithm}
\usepackage{algorithmic}
\usepackage{float}
\usepackage{mathtools}
\usepackage[noabbrev,capitalize]{cleveref}
\usepackage{pdfpages}  
\usepackage{color}
\usepackage{custom_style}
\usepackage{mathrsfs} 
\usepackage{bm}
\usepackage{enumitem}
\usepackage{wrapfig}

\usepackage{lineno}
\usepackage{thm-restate}  
\usepackage{multirow}

\usepackage[accepted]{icml2024}

\theoremstyle{plain}
\newtheorem{theorem}{Theorem}[section]

\newtheorem{lemma}[theorem]{Lemma}

\theoremstyle{definition}
\newtheorem{definition}[theorem]{Definition}

\theoremstyle{remark}
\newtheorem{remark}[theorem]{Remark}

\newcommand{\dd}{\mathrm{d}}
\newcommand{\PP}{\mathbb{P}}

\newcommand{\eps}{\varepsilon}

\graphicspath{{./figure/}}

\begin{document}

\icmltitlerunning{Generative Conditional Distributions by Neural (Entropic) Optimal Transport}

\twocolumn[
\icmltitle{Generative Conditional Distributions by Neural (Entropic) Optimal Transport}





\begin{icmlauthorlist}
\icmlauthor{Bao Nguyen}{cuhk,vinuni}
\icmlauthor{Binh Nguyen}{nus}
\icmlauthor{Hieu Trung Nguyen}{cuhk,vinai}
\icmlauthor{Viet Anh Nguyen}{cuhk}

\end{icmlauthorlist}

\icmlaffiliation{cuhk}{The Chinese University of Hong Kong}
\icmlaffiliation{vinuni}{VinUniversity}
\icmlaffiliation{nus}{Department of Mathematics, National University of Singapore}
\icmlaffiliation{vinai}{VinAI Research}

\icmlcorrespondingauthor{Bao Nguyen}{bao.nn2@vinuni.edu.vn}

\icmlkeywords{Conditional Distribution, Optimal Transport, Generative Models}

\vskip 0.3in
]



\printAffiliationsAndNotice{}  
\begin{abstract}
  Learning conditional distributions is challenging because the desired outcome is not a single distribution but multiple distributions that correspond to multiple instances of the covariates. We introduce a novel neural entropic optimal transport method designed to effectively learn generative models of conditional distributions, particularly in scenarios characterized by limited sample sizes. Our method relies on the minimax training of two neural networks: a generative network parametrizing the inverse cumulative distribution functions of the conditional distributions and another network parametrizing the conditional Kantorovich potential. To prevent overfitting, we regularize the objective function by penalizing the Lipschitz constant of the network output. Our experiments on real-world datasets show the effectiveness of our algorithm compared to state-of-the-art conditional distribution learning techniques. Our implementation can be found at \url{https://github.com/nguyenngocbaocmt02/GENTLE}.
\end{abstract}

\section{Introduction}

The conditional distribution of a response variable given the covariate information is a key quantity for numerous tasks in machine learning and data science. For example, knowing the conditional distribution of a patient's health outcome (the response) given the patient's characteristics (the covariate) improves risk assessment and prediction accuracy and allows personalized interventions wherein healthcare providers can tailor treatments based on individual patient profiles. It is also a useful tool to evaluate the effects of a policy in economics \cite{ref:diaz2020machine, ref:athey2015machine}. Access to a response's conditional probability distribution also leads to an estimate of the treatment effect without explicit, additional experiments. 

Unfortunately, learning the relationship between the response and the covariate is challenging, especially in high-dimensional covariate space or complex dependency settings. Many simple and interpretable methods, such as kernel density estimation, may struggle to capture the complexity of the data-generating conditional distribution accurately~\cite{ref:athey2021using}. On the other hand, there are overly complex (deep neural network) models that perform well on the training data but generalize poorly to new data; this overfitting behavior triggers unreliable predictions and interpretations. Additionally, performing inferences on these overly complex models can be time-consuming and environmentally unfriendly. Alternatively, using additional expert inputs, one can build physically informed simulation models to represent the real physical world. However, running simulations can be time-consuming, making them ill-suited for real-time decision-making.

\begin{figure}[h]
    \centering
    \includegraphics[width=\linewidth]{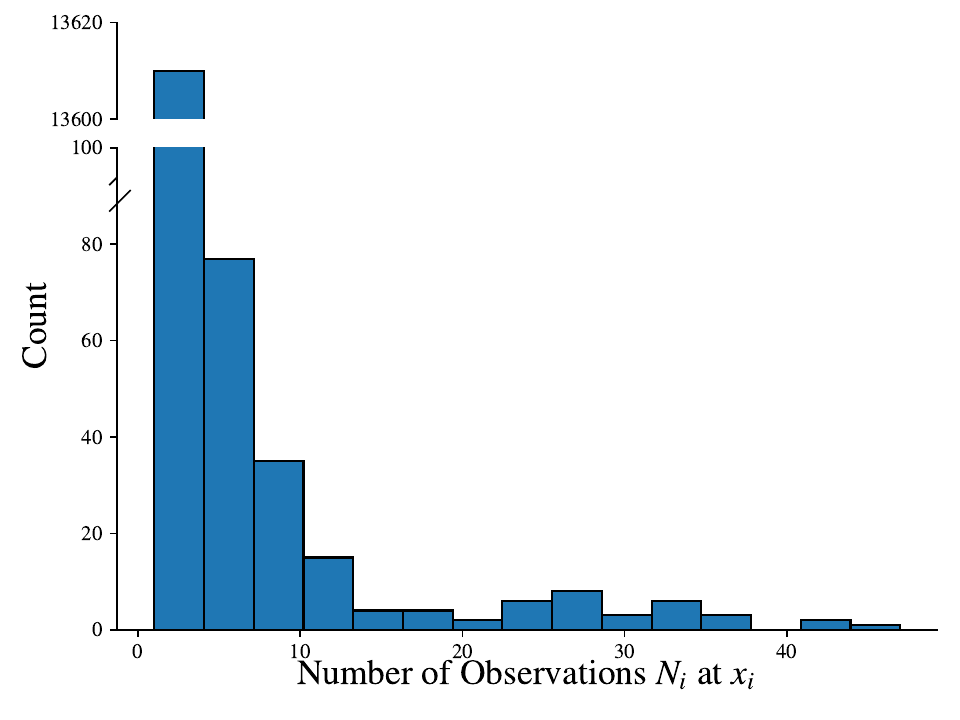}
    \vspace{-7mm}
    \caption{Histogram count of Number of observed responses for distinct covariate values in the LDW dataset~\cite{ref:lalonde1986evaluating}. This dataset has over 13,000 covariates with only one response.}
    \label{fig:count-unique}
\end{figure}

Another critical obstacle underlying conditional distribution learning is that we are not simply learning one distribution; instead, we need to learn a collection of distributions, one conditional distribution for each possible value of the covariate information. This challenge intensifies because, in reality, the data is scarce. This scarcity is particularly prevalent in healthcare, where privacy concerns or budget constraints prohibit the collection and dissemination of samples. As an illustrative example, we will consider the Lalonde-Dehjia-Wahba (LDW) \citep{ref:lalonde1986evaluating, ref:dehejia1999causal} dataset, which is a widely used dataset for analyzing the average treatment effects~\citep{ref:athey2021using}. The LDW dataset comprises 16,177 samples aggregated from actual experiments and the Current Population Survey. Each observed sample includes nine covariate attributes of a person and a scalar response reporting the individual's earnings in 1978. Since two persons may have the exact attributes, it happens that for a specific covariate value, there are multiple observations of the response. The histogram in Figure~\ref{fig:count-unique} shows the count numbers per distinct covariates for the LDW dataset. We observe that more than 13,000 covariates have only one response. This is an extreme case where we must learn the conditional distribution with a single data point observed at that covariate. On the other hand, we also observe several popular covariates with many responses: the most popular covariate has 47 responses. This again induces unbalanced sampling in which the number of responses is not balanced across different covariate values. A naive solution to fix this problem could be under-sampling (remove data) the more frequent covariate and over-sampling (create synthetic data) for the rare covariate to balance the dataset using popular methods such as SMOTE \cite{ref:chawla2002smote}, SMOTER~\cite{ref:torgo2013smote} and SMOGN~\cite{ref:branco2017smogn}. Unfortunately, these methods do not scale well for large datasets with more than thousands of covariates.

The current literature on conditional distribution learning reveals a notable gap in methodologies that can effectively address the aforementioned difficulties. Specifically, our experimental section shows that CWGAN~\cite{ref:athey2021using} suffers from mode collapse and fails to recover the data-generating conditional distributions.

\subsection{Problem Statement and Contributions}

We consider $X$ a multi-dimensional covariate and $Y$ a one-dimensional response. For any specific value $x$, the data-generating conditional distribution of $Y| X=x$ is denoted by $\mu_x$. Our goal is to learn these conditional distributions $\mu_x$ uniformly over all $x$ from a finite dataset $\cD = \{ (x_i, \{y_{i,k}\}_{k=1, \ldots, N_i} ) \in \mathbb{R}^d \times \mathbb{R}$. Without any loss of generality, we suppose that the covariates $x_i$ are distinct, and for each $x_i$, $\{y_{ik}\}_{k=1, \ldots, N_i}$ are $N_i$ independent samples drawn from $\mu_{x_i}$.


More importantly, we aim to learn a \textit{generative} model of the conditional distributions, which can allow us to draw new, unseen samples. This task is more challenging than learning specific finite-dimensional statistics of the conditional distributions, such as learning conditional expectations or quantiles, especially under a small training sample size. We reemphasize that conditional distribution learning is challenging: our goal is not simply learning one distribution, but instead, a collection of distributions -- one conditional distribution for each possible value of the covariate information. This challenge intensifies because of data scarcity. This difficulty is often encountered in real-world deployment, for instance, in health care and finance.

\textbf{Contributions.} We introduce GENTLE, a \textbf{ge}nerative \textbf{n}eural \textbf{t}ransport \textbf{le}arning model for conditional distributions. GENTLE is represented by a neural network $T_\theta$ that takes the covariate $x$ and a uniform $(0, 1)$ random variable $U$ as input, and it outputs $T_\theta(x, U)$ with a distribution that is close to $\mu_x$, the data-generating conditional distribution of $Y|X = x$. Due to the low sample size, $T_\theta$ can overfit, and we propose learning $\theta$ by minimizing the sum of the fitness and the overfitting regularization trade-off. The overfitting regularizer ensures that $T_\theta$ is sufficiently smooth in the covariate $x$, and we use an entropic optimal transport distance to measure the discrepancy of the generative distribution outputs. To this end, we create a second network $v_\phi$ to approximate the conditional potential associated with the semi-dual formulation of the entropic optimal transport distance. This leads to a min-max optimization problem over the pair of generative-conditional potential network parameters $(\theta, \phi)$. This min-max problem is then solved using a state-of-the-art gradient descent-ascent algorithm.

Using real-life datasets, we empirically validate that GENTLE performs better than state-of-the-art methods such as conditional Wasserstein GANs across multiple performance metrics (Wasserstein metric and Kolmogorov-Smirnov metric). In particular, even for the case of LDW-CPS data that exhibits severe imbalances in observed responses, GENTLE still demonstrates a remarkable ability to generate observations that follow ground truth distributions closely, quantitatively, and qualitatively.

This paper unfolds as follows: Section~\ref{sec:literature} provides an overview of the existing literature. Section~\ref{sec:background} reviews the technical background on optimal transports and the tools we will utilize. Section~\ref{sec:method} presents our approach to generative conditional distributions. The numerical results are provided in Section~\ref{sec:experiment}.

\textbf{Notations.} We write the capital letter $X$ for the random variable and the case letter $x$ for a specific realization of $X$. For simplicity, we use $Y(x)$ to denote the conditional random variable $Y | X = x$. For a measurable map $f: \mathbb{R} \to \mathbb{R}$ and any probability measure $\PP$ on $\mathbb{R}$, we use $f_\# \PP$ to denote the pushforward measure of $\PP$ under $f$. 
The (standard) uniform distribution on $(0, 1)$ is denoted $\cU(0, 1)$, and we use $\mathbb{U}$ to represent the probability measure associated with this uniform distribution. We denote $N = \sum_{i} N_i$ as the total number of observations in the training dataset. 

\section{Related Works} \label{sec:literature}
\textbf{Metamodeling} is a computational approach that efficiently captures the relationship between input variables and key output statistics like means or quantiles. This is crucial for real-time simulation applications such as personalized decision making~\cite{ref:shen2021ranking} and online risk monitoring~\cite{ref:jiang2020online}. Traditional metamodeling techniques, while effective, often require pre-specification of relevant statistics, limiting their flexibility. To address this, generative metamodeling has emerged as a promising technique that estimates conditional distributions quickly and dynamically. Generative metamodeling differs from traditional approaches by simulating a wider range of possible outcomes, providing a more comprehensive analysis of potential scenarios. \citet{ref:athey2021using} propose using a Wasserstein Generative Adversarial Model (WGAN, \citealt{ref:arjovsky2017wasserstein}) to simulate the distribution of outcomes conditioned on covariates, offering a novel approach to outcome simulation. Similarly, \citet{ref:hong2023learning} propose quantile-regression-based generative metamodeling (QRGMM), another method of conditional distribution learning. QRGMM leverages the conditional quantile regression technique to provide a more detailed understanding of the distribution tails, which is crucial in risk assessment and decision-making processes. These advancements in generative metamodeling can significantly enhance the accuracy and applicability of real-time simulations, paving the way for more personalized and precise decision-making tools in various fields.

\textbf{Synthetic evaluation data.} Previous works~\cite{ref:van2024can, ref:parikh2022validating} leverage the deep generative model to approximate conditional distribution for synthetic data generation. \cite{ref:shrivastava2019learning} parameterize a conditionally Gaussian model with a Deep Neural Network to model error distribution for each state of the object.

\textbf{Neural Optimal Transport} is a recent emergent framework that proposes to learn the optimal transport plan by deep neural networks. Various existing works in this direction include parameterization of the OT map as the gradient of an input convex neural network \cite{ref:jacob2018w2gan,ref:makkuva2020optimal,ref:bunne2022supervised,ref:bunne2023learning}; or solutions of convex-concave problems \cite{ref:korotin2022neural,ref:gushchin2022entropic,ref:choi2023generative}. However, these formulations are primarily suited to computer vision or computational biology applications. To the best of our knowledge, our paper is the first work attempting to formulate the neural OT framework into a conditional distribution learning task.

\textbf{Learning and decision-making with conditional distributions} also attracted tremendous attention in artificial intelligence and operations research communities. \citet{ref:nguyen2020distributionally} studied the robust parameter estimation of conditional distributions under the parametric exponential family of distributions, while \citet{ref:nguyen2020distributionally2} studied the non-parametric Wasserstein robust estimation of conditional statistics. Making conditional decisions based on observed covariates is a well-known problem in finance~\citep{ref:nguyen2024robustifying} and operations research~\citep{ref:esteban2022distributionally}; interested readers are invited to the survey by~\citet{ref:sadana2024survey} for recent developments.

\section{Background} \label{sec:background}

We focus on a one-dimensional response space for $Y$. The optimal transport distance between two distributions supported on $\mathbb{R}$ is formally defined as follows.

\begin{definition}[Optimal Transport Distance, \citealt{ref:villani2009optimal}]
Given two probability measures $\mu$ and $\nu$ supported on $\mathbb{R}$ with finite second moments, the optimal transport (2-Wasserstein) distance between them is defined as
\begin{equation}
    \label{eq:w2-distance}
    \mathrm{W}_2^2(\mu, \nu) \egaldef\inf_{\pi\in\Pi(\mu,\nu)}\int_{\mathbb{R}\times \bbR} |y -y' |^2 \, \mathrm{d}\pi(y, y'),
\end{equation}
where $\Pi(\mu, \nu)$ denotes the set of probability distributions on $\mathbb{R}^2$ with marginals $\mu$ and $\nu$, respectively.
\end{definition}

\paragraph{Optimal transport in 1d.}

In one dimension, the 2-Wasserstein distance admits a closed form expression \citep[Theorem~2.10]{ref:bobkov2019one}: for two measures $\mu$ and $\nu$ with cumulative density functions (CDF) $F_{\mu}$ and $F_{\nu}$, respectively, we have
\begin{equation} \label{eq:wasserstein}
    \begin{aligned}
        \mathrm{W}_2^2(\mu, \nu) 
        &= \int_{0}^1 | F_{\mu}^{-1}(t) - F_{\nu}^{-1}(t) |^2 \dd t \\
        &= \mathbb{E}_{U \sim \cU(0,1)}\left[ | F_{\mu}^{-1}(U) - F_{\nu}^{-1}(U)|^2 \right].
    \end{aligned}
\end{equation}

\paragraph{Entropic-regularized optimal transport.}
The recent popularity in applications of OT to machine learning is largely due to its efficient computation by adding negative entropy of the transport plan as a form of regularization to the linear OT program~\eqref{eq:w2-distance}. This problem is called entropic optimal transport (EOT), which is smooth and strongly convex, therefore has a unique minimizer:
\begin{equation}
    \label{eq:w2-eot}
    \mathrm{W}_{2,\varepsilon}^2(\mu, \nu) \egaldef \min_{\pi \in \Pi(\mu, \nu)} \int_{\mathbb{R}^2} |y - y'|^2 \dd \pi - \varepsilon H(\pi). 
\end{equation}
\citet{ref:cuturi2013sinkhorn} showed that the minimization problem~\eqref{eq:w2-eot} can be solved by Sinkhorn’s algorithm (a type of matrix scaling algorithm, also called iterative proportional fitting procedure, \citealt{ref:kullback1968probability}) that can be implemented at large scale and is analytically tractable. In addition to its computational advantages, problem~\eqref{eq:w2-eot} can also be interpreted as a specific instance of the Schrödinger bridge problem, which has a rich history in physics. 
For an in-depth theoretical introduction to EOT, readers can refer to \citet{ref:nutz2021introduction} or \citet{ref:leonard2014survey}. 

A nice property of the EOT problem~\eqref{eq:w2-eot} is that it allows an equivalent smooth unbounded semi-dual formulation~\citep[Proposition 2.1]{ref:genevay2016stochastic}
\begin{equation}
    \label{eq:eot-semidual}
    \begin{aligned}
    &\min_{\pi \in \Pi(\mu, \nu)} \int_{\mathbb{R}^2} |y - y'|^2 \dd \pi - \varepsilon H(\pi) \\
    &= \sup_{v(\cdot) \in \mathcal{C}} \int_{\mathbb{R}} v^{
    \varepsilon}(y) \dd \mu + \int_{\mathbb{R}} v(y') \dd \nu - \varepsilon,
    \end{aligned}
\end{equation}
where $\mathcal C$ is the space of all continuous functions from $\mathbb{R}$ to $\bbR$, and $v^{\eps}$ is defined as
\begin{equation}\label{eq:soft-c-transform}
    v^{\eps}(y) = - \varepsilon\log \left(\int_{\mathbb{R}} \exp\left(\frac{v( y') - \abs{y - y'}^2 ) }{\varepsilon}\right) \dd \nu \right).
\end{equation}
The function $v$ is called Kantorovich potential, along with its smoothed $c$-transform $v^{\eps}$, with $c$ being the absolute difference bi-function. Using the \textit{semi}-dual form is that we need to optimize over only one function $v$ associating with $\nu$, and we can obtain $v^\eps$ as the potential associating with $\mu$.

\section{Methodologies} \label{sec:method}

We propose to learn a mapping $T: \bbR^d \times (0, 1) \to \bbR$ such that uniformly over all $x$, $T(x, U) = Y(x)$ in distribution when $U \sim \cU(0, 1)$. The existence of this mapping $T(x, U)$ is guaranteed by the following celebrated result.

\begin{lemma}[Noise Outsourcing, \citealt{ref:austin2015exchangeable}]
    Suppose that $\mathcal X$ and $\mathcal Y$ are standard Borel spaces and that $(X, Y)$ is an $(\mathcal X \times \mathcal Y)$-valued random variable. Then, there are random variables $U \sim \cU(0, 1)$ coupled with $\mathcal X$ and $\mathcal Y$ and a Borel function $T : \mathcal X \times (0, 1) \rightarrow \mathcal Y$ such that $U$ is independent from $X$ and
    \[
        (X, Y) = (X, T(X, U)) \quad \text{almost surely}.
    \]
\end{lemma}
In general, the mapping $T$ is not unique. Following the neural OT framework~\cite{ref:korotin2022neural,ref:makkuva2020optimal,ref:bunne2022supervised}, we use a neural network to parametrize a possible instance of $T(x, U)$, and we denote this network as $T_\theta(x, U)$. For any specific value of the parameter $\theta$, the network outputs a probability measure $T_\theta(x, \cdot)_\# \mathbb{U}$, which represents the distribution of $T_\theta(x, U)$ with $U \sim \cU(0, 1)$. We propose the following regularized loss to find the optimal map $T_\theta$:
\begin{equation}
\label{eq:loss-regularize}
\min_\theta \mathrm{Fit}(\theta) + \lambda \mathrm{Reg}(\theta),
\end{equation}
where the first term is a measure of the fitness of the map, the second is a regularization to prevent overfitting, and $\lambda \ge 0$ is the regularization weight. We explain the detailed formulation of each term and its rationale in the next sections.

\subsection{Fitness Measure} \label{sec:fit}
Motivated by the OT distance in~\eqref{eq:wasserstein}, a natural choice to measure the discrepancy between the network outputs and the conditional data-generating distributions is by integrating the discrepancy of the respective inverse CDFs over all possible values $x$, \ie, by evaluating
\begin{equation}
    \begin{aligned}
& \bbE_{X \sim \cD, U \sim \cU(0, 1)} [| F_{\mu_{X}}^{-1}(U) - T_\theta(X, U) |^2 ] \\
=& \mathbb{E}_{X \sim \mathcal D} \left[ \mathbb{E}_{U \sim \cU(0, 1)} [| F_{\mu_X}^{-1}(U) - T_\theta(X, U) |^2]\right].
\end{aligned}
\end{equation}
Above, $F_{\mu_x}$ is the cumulative distribution function of $Y(x)$, and the outer expectation is taken over the (uniform) sampling of all distinct values of $X$ in the training dataset $\mathcal D$. Unfortunately, the inverse CDF $F_{\mu_X}^{-1}$ can be unbounded, and the magnitude of $F_{\mu_X}^{-1}$ explodes to infinity for tail events whenever $U$ is to close to the boundary 0 or 1. Hence, learning $T_\theta$ with the above metric is not stable. To improve stability, we use the following fitness function
\begin{equation}
    \label{eq:fit}
    \mathrm{Fit}(\theta) \egaldef \mathbb{E}_{X \sim \mathcal D} \big[ \mathbb{E}_{U \sim \cU(0, 1)} [| U - F_{\mu_X}(T_\theta(X, U)) |^2]\big].
\end{equation}

In practice, we do not have access to the real cumulative distribution of $Y(x)$. However, we have a set of responses sampled from $Y(x)$, so we propose to use a simple method such as Kernel Density Estimator (KDE) with the kernel bandwidth of $h$ to have a smooth conditional density from responses of the covariate $x$ in the training dataset. The kernel bandwidth plays a vital role in our objective function, and tuning this parameter is important. In theory, our KDE smoothness using the Gaussian kernel is strongly related to the smoothed Wasserstein distance. From~\citet[Lemma 1 and Corollary 1]{nietert2021smooth}, the smooth-2-Wasserstein distance is continuous and monotonically non-increasing in the Gaussian kernel bandwidth. This means that the larger the bandwidth (the more smoothing applied), the smaller the fitness term will become. In practice, we could implement an automatic kernel bandwidth selection such as the Scott~\citep{scott1979optimal} or the Silverman~\citep{silverman2018density} methods. In our implementation, the value of the bandwidth is selected by a grid search to minimize the Wasserstein distance on the validation set.

\subsection{Regularization Measure}
\label{sec:reg_measure}
\begin{figure}
    \centering
    \includegraphics[width=\linewidth]{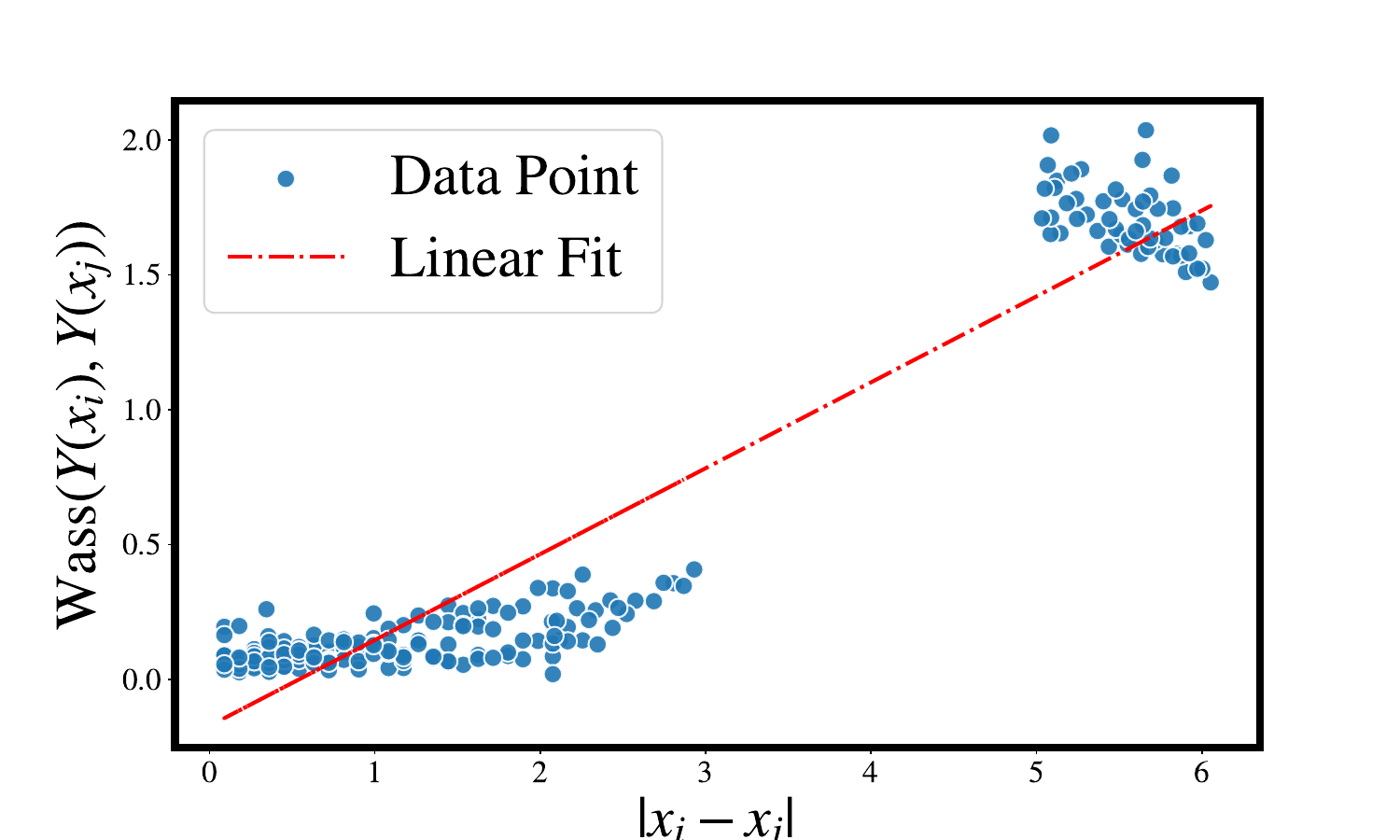}
    \caption{Positive correlation between the Wasserstein distance between $Y(x_i)$ and $Y(x_j)$ and the covariate distance $\| x_i - x_j \|$. Only covariates $x$ with more than 18 observations are selected.}
    \label{fig:lipschitz}
\end{figure}

Minimizing the $\mathrm{Fit}(\theta)$ term can lead to overfitting due to the low sample sizes for many values of $x_i$. To combat this problem, we add a regularization term to promote the transfer learning across different values of $x_i$. Consider the CPS dataset, we examine the relationship between the covariate distance $\| x_i - x_j \|$ versus the empirical Wasserstein distance between the conditional random variables $Y(x_i)$ and $Y(x_j)$. For better accuracy, we select only pairs $(x_i, x_j)$ such that both $N_i$ and $N_j$ are greater than 18. The scatter plot in Figure~\ref{fig:lipschitz} indicates a positive correlation between the covariate distance and the Wasserstein distance of the conditional distributions. We remark that we observed the same trend in other datasets.  

From this empirical evidence, we can assume that if two covariates $x_i$ and $x_j$ are similar, their respective conditional distributions of $Y(x_i)$ and $Y(x_j)$ should also be similar. We emphasize that relying on this assumption to design the regularization term does not restrict the generalization of our framework. This assumption, in general, holds for the case where the data-generating process follows the form $Y = f(X) + \text{noise}$, where $f$ is a continuous function. This is a classic setting in distribution learning, a more general form of linear assumption. Moreover, classic algorithms for supervised and unsupervised learning, such as kNN or k-clustering, also require this assumption to work. We can, therefore, assume a Lipschitz condition that there exists a finite positive number $L$ such that
\[
    W_2^2( \mu_{x_i}, \mu_{x_j} ) \le L \| x_i - x_j \| \qquad \forall i, j,
\]
where $\mu_x$ is the data-generating conditional probability measure for $Y(x)$. 
If the generative network $T_\theta$ matches sufficiently well the conditional distributions, then it is reasonable to translate the above Lipschitz assumption into a regularization term: let $\mathcal E$ be a set of pairs $(x_i, x_j)$ for which we would like to impose Lipschitz conditions, we can regularize using the sum of the pairwise \textit{entropic} OT distance:
\begin{equation} \label{eq:eot}
\sum_{(x_i, x_j) \in \mathcal E} \mathrm{W}^2_{2,\varepsilon}\left(T_{\theta}(x_i, \cdot)_\# \bbU, T_{\theta}(x_j, \cdot)_\# \bbU \right).
\end{equation}
The semi-dual form of the EOT~\eqref{eq:eot-semidual} leads to
\begin{align*}
        \sum_{(x_i, x_j) \in \mathcal E}~\Big( \max_{v_{ij}(\cdot) \in \mathcal{C}} \int_{\mathbb{R}} v_{ij}^{\varepsilon}(y) \dd \PP_{\theta, j} + \int_{\mathbb{R}} v_{ij}(y) \dd \PP_{\theta,i} - \varepsilon \Big).
\end{align*}
Above, we have used the shorthand $\mathbb{P}_{\theta,i} = T_\theta(x_i, \cdot)_\# \mathbb{U}$, the pushforward measure of the uniform measure under the map $T_\theta(x_i, \cdot)$. Each maximization problem inside the sum is an infinite-dimensional optimization problem, in which we need to search over all potentials $v_{ij}$ continuous. To alleviate this computational burden, a naive approach is to use a neural network $v_\phi(x_i, x_j, \cdot) \approx v_{ij}(\cdot)$, where $\phi$ is a learnable parameter. By omitting the constant $\eps$ and switching the order of the summation and the maximization operator, we can regularize using 
\begin{align*}
    \max_\phi\!\!\!\sum_{(x_i, x_j) \in \mathcal E}\!\!\!\left( \int_{\mathbb{R}} v_{\phi}^{\varepsilon}(x_i, x_j, y) \dd \PP_{\theta, j}\!+\!\int_{\mathbb{R}} v_\phi(x_i, x_j, y) \dd \PP_{\theta,i} \right).
\end{align*}
A downside of this parametrization is that it requires both $x_i$ and $x_j$ as input to $v_\phi$ to output the potential. This leads to redundancy in the parametrization. To combat this problem, we propose a \textit{single} input network $v_\phi(x_i, y)$ to approximate $v_{ij}(y)$, the potential associating with $\mathbb{P}_{\theta, i}$. Further, we compute the smoothed transform $v_\phi^\eps(x_i, y)$
\begin{align*}
& v_\phi^{\eps}(x_i, y) \egaldef \\
&- \varepsilon\log \left(\int_{\mathbb{R}} \exp\left(\frac{v_\phi( x_i, y') - \abs{y - y'}^2 ) }{\varepsilon}\right) \dd \PP_{\theta, i} \right),
\end{align*}
and use $v_\phi^{\eps}(x_i, y)$ to approximate $v_{ij}^\eps(y)$, the potential associating with $\mathbb{P}_{\theta, j}$. The above single input parametrization no longer takes $x_j$ as input; thus, to make the network well-defined, we need disambiguation by imposing the following condition: for each distinct value $x_i$, there is \textit{at most one} pair $(x_i, x_j)$ that belongs to~$\mathcal E$. If this condition on $\mathcal E$ holds, then we can uniquely identify the $x_j$ such that $v_\phi^{\eps}(x_i, y)$ is serving as the potential.

Thus, for a valid set $\mathcal E$, we propose the following regularization term:
\begin{subequations}
\label{eq:reg}
\begin{equation}
\mathrm{Reg}(\theta) \egaldef \max_\phi~\mathcal R(\theta, \phi),
\end{equation}
where $\mathcal R(\theta, \phi)$ is
\begin{align}
& \mathcal R(\theta, \phi) \egaldef  \\
&\quad \sum_{(x_i, x_j) \in \mathcal E}~\Big( \int_{\mathbb{R}} v_{\phi}^{\varepsilon}(x_i, y) \dd \PP_{\theta, j} + \int_{\mathbb{R}} v_\phi(x_i, y) \dd \PP_{\theta,i} \Big). \notag 
\end{align}
\end{subequations}
In this way, $\mathrm{Reg}(\theta)$ is the sum of the semi-dual neural entropic OT distances of all pairwise distributions specified by $\mathcal E$.

To conclude this section, we elaborate on a reasonable approach to building $\mathcal E$. Ideally, $\mathcal E$ should capture neighborhood information about which pair $(x_i, x_j)$ is close to each other; further, it should satisfy our disambiguation criterion that for each distinct value $x_i$, there is \textit{at most one} pair $(x_i, x_j)$ that belongs to~$\mathcal E$. A reasonable choice of $\mathcal E$ that is also computationally efficient is to build $\mathcal E$ as the \textit{directed} edge set of a minimum spanning tree of all $x$. Here, the spanning tree minimizes the sum of distance $\| x_i - x_j \|$ between two connected nodes, and we can employ Kruskal's algorithm \citep{ref:kruskal1956shortest}. For the direction of the edge: at each iteration, when we add $x_i$ to the incumbent minimum spanning tree, then $x_i$ is chosen as the tail of the edge, and the other endpoint of the adding edge (which already belongs to the incumbent tree) is chosen as the head of the edge. This construction will satisfy the disambiguation criterion for the set $\mathcal E$. We emphasize that the graph-based approach to identify which $x_j$ is associated with $x_i$ in the regularization term is not restricted only to a discrete set of covariates. This is because the graph-based approach is only employed during the training when the number of covariates is finite. In inference phases, the model can give the distributional prediction to any covariates.

\begin{remark}
    In~\eqref{eq:fit}, we use the unregularized OT distance as a discrepancy function for the fit term $\mathrm{Fit}(\theta)$ because therein, we are measuring the discrepancy between a fixed target (a uniform distribution $U$) and a quantity that depends on the generative network output $T_\theta(X, U)$ and the data-generating CDFs $F_{\mu_X}$. This contrasts with the regularization term $\mathrm{Reg}(\theta)$, where we need to compare two distributions that are both outputs of the generative network $T_\theta$. Using the unregularized OT distance for the $\mathrm{Reg}(\theta)$ will require evaluating the CDFs of $T_\theta$. To alleviate this difficulty, we use the semi-dual form of EOT, where we can parametrize a single conditional potential network to evaluate all the pairwise EOT distances between covariate pairs in $\mathcal E$.     
\end{remark}

\subsection{Min-Max Optimization with Smooth Gradient Descent-Ascent}

By substituting the form of the regularization term~\eqref{eq:reg} into~\eqref{eq:loss-regularize}, we obtain the min-max optimization problem:
\begin{equation} \label{eq:minmax-loss}
\min_\theta \max_\phi~\mathrm{Fit}(\theta) + \lambda \mathcal R(\theta, \phi).
\end{equation}
Naively optimizing problem~\eqref{eq:minmax-loss} will result in a common pitfall in optimization of training GANs: gradient descent-ascent (GDA) on nonconvex-nonconcave objective is, in general, hard to converge \cite{ref:heusel2017gans,ref:lin2020gradient,ref:zheng2023universal}.
To alleviate this problem, we employ the state-of-the-art first-order algorithm for GDA \cite{ref:zheng2023universal} with a convergence guarantee.
First, for parameters $r_1 \neq r_2$, $r_1, r_2 > 0$, we add two extra regularization (smoothing) terms with auxiliary variables $p$ and $q$ into the loss function. This leads to the augmented objective
\begin{equation}
\label{eq:final-minmax-loss}
\begin{aligned}
    &\cL(\theta, \phi, p, q) \\
    &\egaldef \mathrm{Fit}(\theta) + \lambda \cR(\theta, \phi) + \frac{r_1}{2} \| \theta - p \|_2^2 - \frac{r_2}{2} \| \phi - q \|_2^2.
\end{aligned}
\end{equation}
Including the quadratic terms enhances the smoothness of the primal and dual update, facilitating a more balanced trade-off between the gradient updates in our nonconvex-nonconcave setting.
In practice, we must evaluate this loss with training samples, detailed in Algorithm~\ref{alg:empirical-loss}.
Our training procedure is summarized in Algorithm~\ref{alg:minimax-training}. 
This doubly-smoothed GDA update is guaranteed to converge to a stationary point \citep[Theorem 2]{ref:zheng2023universal} as long as the max term $\cR(\theta, \phi)$ satisfies Kurdyka-Łojasiewicz condition (a nonsmooth extension of the more popular Polyak-Łojasiewicz condition).

In general, we resort to a validation procedure to choose the appropriate parameters $h, \lambda, \varepsilon, r_1, r_2$, and the optimization algorithm's learning rates $(\alpha, \beta, \gamma, \delta)$. Details on the parameters grid are given in the Appendix~\ref{sec:additional-details}. 

\begin{algorithm}[h]
   \caption{Empirical loss evaluation}
   \label{alg:empirical-loss}
\begin{algorithmic}[1]
   \STATE {\bfseries Input:} data $(x_i, \hat{F}_{x_i})$, network $T_\theta$ and $v_\phi$, auxiliary variables $p, q$, batch size $B$, set of pair covariates $\cE$, regularization parameter $\varepsilon, \lambda, r_1, r_2$
   \STATE {\bfseries Output:} loss estimate $\widehat{\cL}(\theta, \phi, p, q)$
    \STATE {\bfseries Initialize:} take a batch of $B$ training nodes $\{x_i^{(b)}\}_{b\in [B]}$
    \FOR{each $(x_i^{(b)}, x_j^{(b)}) \in \cE$}
    \STATE Sample $U^{(b)} \sim \cU[0, 1]$
    \FOR{$m = 1, \cdots, M$}
        \STATE Sample $U^{(b, m)} \sim \cU[0, 1]$
    \ENDFOR
    \ENDFOR
   \STATE $\widehat{\mathrm{Fit}}(\theta) \gets \frac{1}{B}\sum_b \left[U^{(b)} - \hat{F}_{x_i^{(b)}}(T_\theta(x_i^{(b)}, U^{(b)})) \right]^2_2$ \hfill (following \eqref{eq:fit})

    \STATE Compute following~\eqref{eq:reg}: \begin{multline*}
        \hat{\cR}(\theta, \phi) \gets \frac{1}{M\times B} \sum_{(x_i^{(b)}, x_j^{(b)}, U^{(b,m)})} \\ \left[v_\phi^\varepsilon(x_i^{(b)}, T_\theta(x_j^{(b)}, U^{(b,m)})) + v_\phi(x_i^{(b)}, T_\theta(x_i^{(b)}, U^{(b,m)}))\right]
    \end{multline*}
    \hfill 
    \STATE Compute following~\eqref{eq:final-minmax-loss}:
    \begin{multline*}
    \widehat{\cL}(\theta, \phi, p, q) \\ \gets \widehat{\mathrm{Fit}}(\theta) + \lambda \hat{\cR}(\theta, \phi) + \frac{r_1}{2} \norm{\theta - p}^2_2 - \frac{r_2}{2} \norm{\phi - q}_2^2
    \end{multline*}
    \hfill 
\end{algorithmic}
\end{algorithm}

\begin{algorithm}[h]
   \caption{Training of the Minimax loss \eqref{eq:final-minmax-loss}}
   \label{alg:minimax-training}
\begin{algorithmic}[1]
   \STATE {\bfseries Input:} data $\mathcal D = (x_i, \{y_{ik}\}_{k = 1, \ldots, N_i})$, network $T_\theta$ and $v_\phi$, learning rate $\alpha, \beta$, extrapolation parameters $\gamma,\delta$
    \STATE {\bfseries Initialize: $\theta^0, \phi^0, p^0, q^0$} 
   \FOR{each number of training iteration $t$}
        \STATE $\theta^{t+1} \gets \theta^t - \alpha\nabla_\theta \widehat{\cL}(\theta^t, \phi^t, p^t, q^t)$
        \STATE $\phi^{t+1} \gets \phi^t + \beta\nabla_\phi \widehat{\cL}(\theta^{t+1}, \phi^t, p^t, q^t)$ 
        \STATE $p^{t+1} \gets p^t + \gamma(\theta^{t+1} - p^t)$
        \STATE $q^{t+1} \gets q^t + \delta(\phi^{t+1} - q^t)$
   \ENDFOR
\end{algorithmic}
\end{algorithm}

\section{Experiments} \label{sec:experiment}

\textbf{Datasets.} To benchmark our proposed method GENTLE, we use the following two datasets and the same preprocessing steps as in~\citet{ref:athey2021using}.

The LDW-CPS dataset, constructed by \citet{ref:lalonde1986evaluating, ref:dehejia1999causal}, is widely used in studies of Average Treatment Effects. This dataset contains 16177 samples from the real-world experiment and the Current Population Survey. Each covariate $x_i$ includes the individual's eight attributes: two earnings measure (\texttt{earning 74}, \texttt{earning 75}), two indicators for ethnicity (\texttt{black}, \texttt{hispanic}), marital status (\texttt{married}),  age (\texttt{age}), two education measures (\texttt{education}, \texttt{nodegree}), and a binary variable for whether this individual receives treatment (\texttt{is\_treated}). The response $y_i$ is \texttt{earning 78}, this individual's yearly income in 1978. This dataset has limited samples for each covariate, shown in Figure~\ref{fig:count-unique}. To achieve better testing evaluation and model selection, we choose covariates based on their frequency for the test and validation sets. Specifically, covariates with a frequency higher than 30 are included in the test set, while those with a frequency higher than 20 but less than or equal to 30 are included in the validation set. According to this split strategy, the training set contains 15,255 samples corresponding to 13,745 covariates, the validation set contains 488 samples corresponding to 19 covariates, and the test set contains 434 samples corresponding to 20 covariates.
    
We also apply frameworks to a practical simulator called the Esophageal Cancer Markov Chain Simulation Model (ECM).\footnote{Code is publicly available at~\url{https://simopt.github.io/ECSim}} This simulator enables us to simulate patients' quality-adjusted life years (QALYs) from their current treatment until death. The approach to generate data for training and testing follows the authors in \citet{ref:hong2023learning}. Each covariate $x_i$ consists of five attributes which are \texttt{Barrett}, \texttt{aspirinEffect}, \texttt{statinEffect}, \texttt{drugIndex}, and \texttt{initialAge}. The response feature for this dataset is \texttt{QALY} (quality-adjusted life years). Because we have access to the simulator, we can sample infinite responses for each covariate. We follow the dataset generation in \citet{ref:hong2023learning} in which they use the simulator to uniformly generate $2 \times 10^4$ different covariates for the train set, and each covariate has one response. The validation set has $200$ covariates that are different from those in the train set, while the test set also has $200$ covariates that are different from both the train and validation sets. Each covariate has $10^4$ responses.

\textbf{Baselines.} We compare our method against state-of-the-art baselines, including CWGAN~\citep{ref:athey2021using}, WGAN-GP~\citep{gulrajani2017improved}, MGAN~\citep{baptista2020conditional}, and CDSB~\citep{shi2022conditional}. We keep the architecture of $T_\theta(x, U)$ and $v_\phi$ simple with seven fully connected layers with ReLU activation.   

We denote $D_{\text{train}}$, $D_{\text{val}}$, and $D_{\text{test}}$  as the set of covariates in the training, validation, and test sets, respectively. While the covariates in the training set are used for training the model parameters, the ones in the validation test are utilized to select hyperparameters. For each distinct $x \in D_{\text{test}}$, the set of actual respective responses is denoted as $Y_{GT}(x)$, while we use GENTLE, CWGAN, WGAN-GP, MGAN, and CDSB to generate $K = 10^4$ approximating points for $Y_{\text{GT}}(x)$, denoted as $Y_{\text{GENTLE}}(x)$, $Y_{\text{CWGAN}}(x)$, $Y_{\text{WGAN-GP}}(x)$, $Y_{\text{MGAN}}(x)$, $Y_{\text{CDSB}}(x)$ respectively. We then compute the average \textit{Wasserstein distance} and \textit{Kolmogorov-Smirnov statistic}~\citep{ref:kolmogorov1933sulla} over $D_{\text{test}}$ between $Y_{\text{GT}}(x)$ and the outputs $Y_{\text{GENTLE}}(x)$, $Y_{\text{CWGAN}}(x)$, $Y_{\text{WGAN-GP}}(x)$, $Y_{\text{MGAN}}(x)$, and $Y_{\text{CDSB}}(x)$, as shown in Table~\ref{tab:generalization}.

\begin{table}[h]
\caption{Wasserstein distance and K-S statistic on generated $\hat{Y}(x)$ using different methods vs.~corresponding ground truth. The symbol $\pm$ represents the standard deviation.}
    \label{tab:generalization}
\begin{center}
\begin{small}
\begin{sc}
\begin{tabular}{cccc}
\hline
Dataset & Method & WD($\downarrow$) & KS($\downarrow$) \\ \hline
\multirow{5}{*}{LDW-CPS} & \begin{tabular}[c]{@{}c@{}}GENTLE\\ \end{tabular} & \begin{tabular}[c]{@{}c@{}}\textbf{3767.62}\\ (\textbf{$\pm$ 1337.56})\end{tabular} & \textbf{0.48 $\pm$ 0.05} \\
& CWGAN & \begin{tabular}[c]{@{}c@{}}13371.99 \\ ($\pm$ 3637.66)\end{tabular} & 0.88 $\pm$ 0.09 \\
& MGAN & \begin{tabular}[c]{@{}c@{}}4096.75 \\ ($\pm$ 1004.5)\end{tabular} & 0.91 $\pm$ 0.19 \\
& WGAN-GP & \begin{tabular}[c]{@{}c@{}}7828.66 \\ ($\pm$ 2306.00)\end{tabular} & 0.99 $\pm$ 0.02 \\
 & CDSB & \begin{tabular}[c]{@{}c@{}}4970.66 \\ ($\pm$ 1094.74)\end{tabular} & 0.56 $\pm$ 0.14 \\ \hline
ECM & GENTLE & \textbf{0.97 $\pm$ 0.14} & \textbf{0.21 $\pm$ 0.02} \\
& CWGAN & 4.42  $\pm$ 2.01 & 0.94  $\pm$ 0.05 \\
& MGAN & 1.09  $\pm$ 0.59 & 0.55  $\pm$ 0.13 \\
& WGAN-GP & 1.59  $\pm$ 0.77 & 0.66  $\pm$ 0.27 \\
 & CDSB & 2.04  $\pm$ 0.43 & 0.47  $\pm$ 0.12 \\ \hline
\end{tabular}
\end{sc}
\end{small}
\end{center}
\end{table}

\textbf{Results.} We plot the qualitative results of estimated densities for different covariates $X$ in Figure~\ref{fig:qual1} and~\ref{fig:qual2}. Overall, synthetic observations generated by GENTLE exhibit a strong resemblance to the ground truth distributions in both datasets, while the ones generated by baselines deviate significantly from the data-generating distributions. For each covariate, the distributions generated by CWGAN seem to collapse into the mean of the training data distribution, while those of MGAN and WGAN-GP tend to shrink towards the mean of the conditional data distribution of the covariate. Although CDSB does not suffer from this collapse tendency, it still fails to match the data-generating distributions accurately. This is consistent with the quantitative results in Table~\ref{tab:generalization}, which show that the data generated by GENTLE have not only smaller Wasserstein distance (WD) and Kolmogorov-Smirnov (KS) values but also smaller standard deviations. This suggests GENTLE's superior performance compared to the other baselines.
 
\subsection{Ablation Study on Training Loss Terms}

We explore the impact of removing specific components from the loss function~\eqref{eq:final-minmax-loss} by considering two ablation versions. The first version omits the smoothing term in the loss function $\mathcal L$ by setting $r_1 = r_2 = 0$, while the second version omits the regularization term by setting $\lambda = 0$. When we exclude the regularization term, the model generates uniformly distributed responses in the LDW-CPS dataset, as illustrated in Figure~\ref{fig:no_reg}. We hypothesize that the absence of $\mathrm{Reg}(\theta)$ compromises the model's ability to enforce the Lipschitz condition between the proximity in the covariates and responses space. This absence significantly hampers the method's performance in the dataset with an unbalanced observation of covariates like LDW-CPS. This problem does not exist in the ECM dataset which covariates are uniformly drawn from the simulation, so the performance of this version in ECM just slightly decreases.  On the other hand, removing the smoothing term leads the model to get trapped in bad local optima, learning only simple two-peak patterns in both two datasets, as depicted in Figure~\ref{fig:no_smoothing} and Figure~\ref{fig:ecm_no_smoothing}. This becomes more apparent when examining Table~\ref{tab:ablation}, where the KS statistic of this version surpasses that of the main method. This improvement may be attributed to the model's inclination to match high-density regions exclusively, lacking overall generalization. In summary, the results in Table~\ref{tab:ablation} highlight that both ablation versions exhibit significantly worse performance than the official method, particularly regarding Wasserstein distance. This underscores the importance of the regularization and smoothing terms in maintaining the model's robustness and enhancing its overall performance.

\begin{table}[t]
\caption{Wasserstein distance and K-S statistic on generated $\hat{Y}(x)$ using different ablation versions of GENTLE. The version GENTLE w/o SMO does not use the smoothing term while GENTLE w/o REG does not use the regularization term. The symbol $\pm$ represents the standard deviation.}
\label{tab:ablation}
\begin{center}
\begin{small}
\begin{sc}
\begin{tabular}{cccc}
\hline
Dataset & Method & WD($\downarrow$) & KS($\downarrow$) \\ \hline
\multirow{3}{*}{LDW-CPS} & GENTLE & \begin{tabular}[c]{@{}c@{}}\textbf{3767.62} \\ (\textbf{$\pm$ 1337.56})\end{tabular} & 0.48 $\pm$ 0.05 \\ \cline{2-4} 
 & \begin{tabular}[c]{@{}c@{}}GENTLE w/o  \\ reg \end{tabular} & \begin{tabular}[c]{@{}c@{}}13625.83 \\ ($\pm$ 562.10)\end{tabular} & 0.57 $\pm$ 0.04 \\ \cline{2-4} 
 & \begin{tabular}[c]{@{}c@{}}GENTLE w/o\\ smo \end{tabular} & \begin{tabular}[c]{@{}c@{}}6075.02\\ ($\pm$ 1340.75)\end{tabular} & \textbf{0.41 $\pm$ 0.04} \\ \hline
 \multirow{3}{*}{ECM} & GENTLE & \begin{tabular}[c]{@{}c@{}}\textbf{0.97 $\pm$ 0.14}\end{tabular} & \textbf{0.21 $\pm$ 0.02} \\ \cline{2-4} 
 & \begin{tabular}[c]{@{}c@{}}GENTLE w/o  \\ reg \end{tabular} & \begin{tabular}[c]{@{}c@{}}1.13 $\pm$ 0.47\end{tabular} & 0.29 $\pm$ 0.04 \\ \cline{2-4} 
 & \begin{tabular}[c]{@{}c@{}}GENTLE w/o\\ smo \end{tabular} & \begin{tabular}[c]{@{}c@{}}1.80 $\pm$ 1.28\end{tabular} & 0.30 $\pm$ 0.05 \\ \hline
\end{tabular}
\end{sc}
\end{small}
\end{center}
\end{table}

\section{Conclusions}
This paper introduces a neural generative model for conditional distributions, whose parameters can be learned from limited data. Our method demonstrates remarkable efficacy in generating distributions over response spaces from given inputs. This is exemplified by successfully applying conditional dataset synthesis tasks on LDW-CPS and ECM datasets, where our framework exhibits superior performances.

An exciting extension of the current work could involve exploring the application of our proposed method to dynamic or time-series data. Incorporating temporal dependencies could enhance the model's ability to capture evolving conditional distributions over time, enabling more accurate predictions in dynamic environments. Additionally, extending the setting of our approach beyond one-dimensional responses could further broaden its scope and impact. Finally, exploring interpretability techniques to elucidate the underlying mechanisms driving the learned conditional distributions could enhance understanding of the model's predictions in real-world decision-making scenarios.

\begin{figure*}
     \centering
     \vspace{-3mm}
     \includegraphics[width=0.8\textwidth]{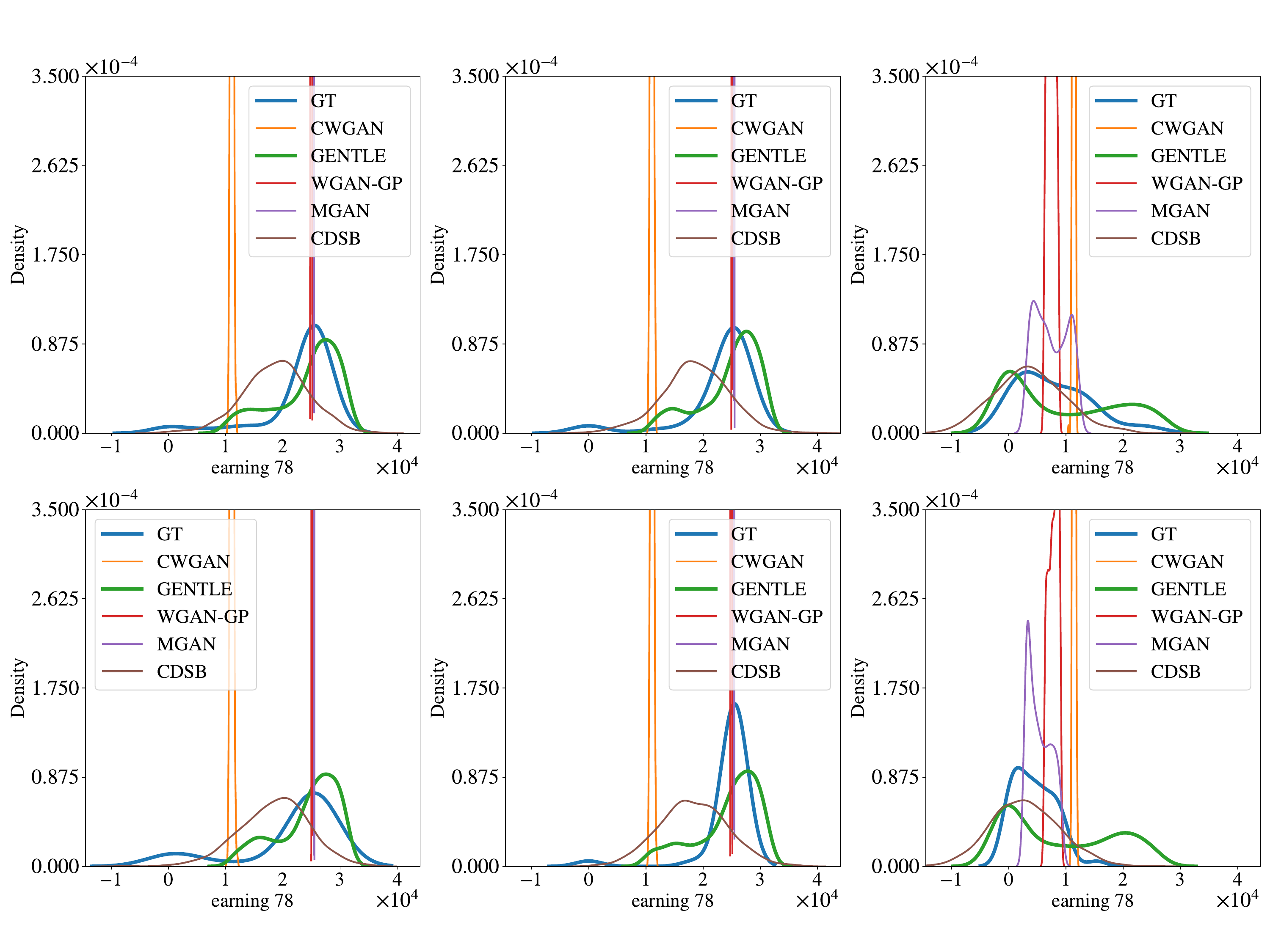}
     \vspace{-7mm}
     \caption{The qualitative results of methods on the LDW-CPS dataset. The density graph `GT', \textbf{`GENTLE' (ours)}, `CWGAN', `WGAN-GP', `MGAN', `CDSB' is constructed by applying kernel density estimate (KDE) on $Y_{\text{GT}}(x)$, $Y_{\text{GENTLE}}(x)$, $Y_{\text{CWGAN}}(x)$, $Y_{\text{WGAN-GP}}(x)$, $Y_{\text{MGAN}}(x)$, $Y_{\text{CDSB}}(x)$, respectively. Each subfigure is for a different value $x$ on the test set. We observe that GENTLE can produce the true distribution much more effectively than baselines.}
     \label{fig:qual1}
 \end{figure*}

\begin{figure*}
     \centering
     \vspace{-3mm}
     \includegraphics[width=0.8\textwidth]{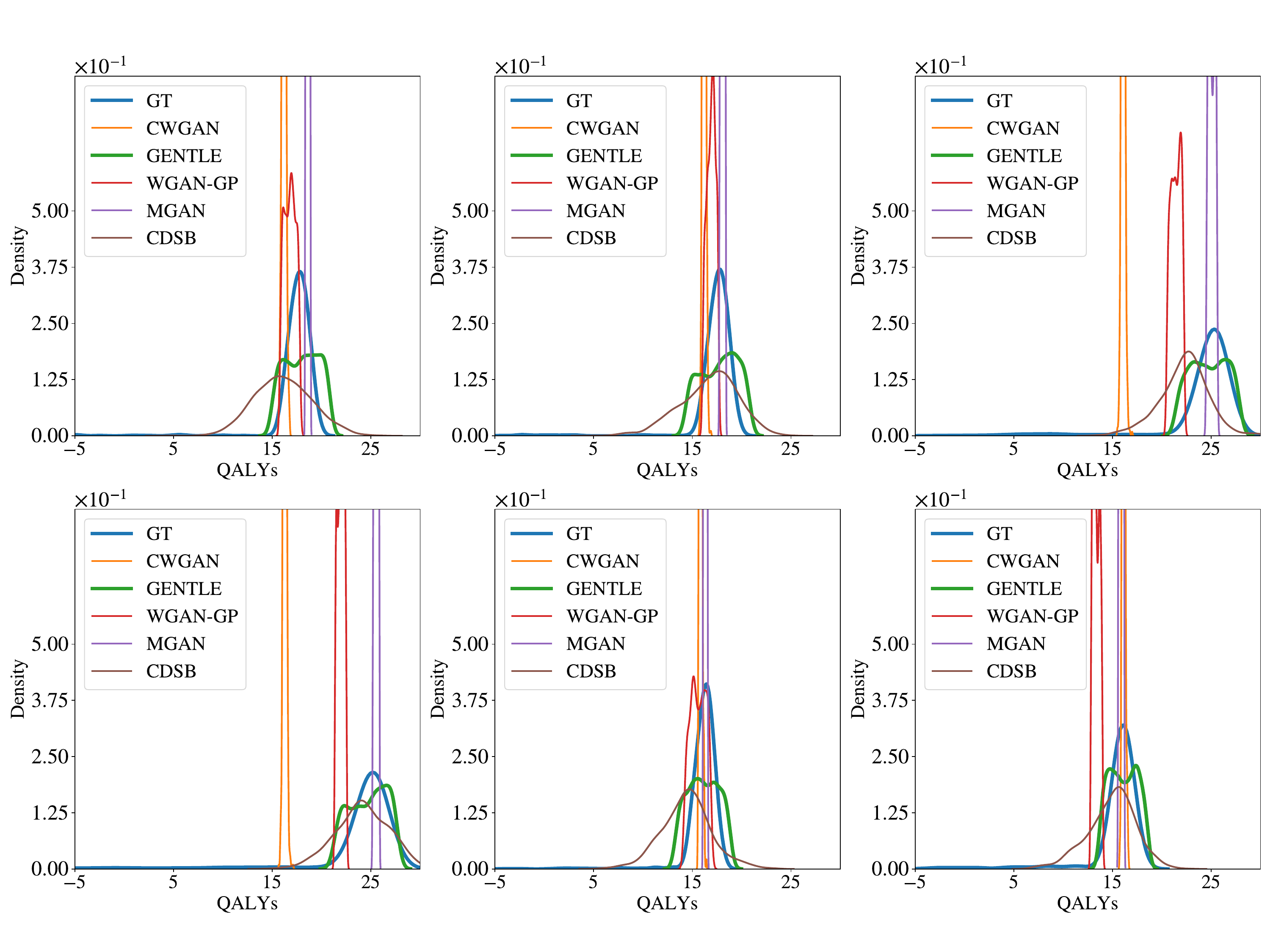}
     \vspace{-7mm}
     \caption{The qualitative results of methods on the ECM dataset. The density graph `GT', \textbf{`GENTLE' (ours)}, `CWGAN', `WGAN-GP', `MGAN', `CDSB' is constructed by applying kernel density estimate (KDE) on $Y_{\text{GT}}(x)$, $Y_{\text{GENTLE}}(x)$, $Y_{\text{CWGAN}}(x)$, $Y_{\text{WGAN-GP}}(x)$, $Y_{\text{MGAN}}(x)$, $Y_{\text{CDSB}}(x)$, respectively. Each subfigure is for a different value $x$ on the test set. We observe that GENTLE can produce the true distribution much more effectively than baselines.}
     \label{fig:qual2}
 \end{figure*}

\textbf{Acknowledgments.} The work of Binh Nguyen is supported by Singapore’s Ministry of Education grant A-0004595-00-00. Viet Anh Nguyen gratefully acknowledges the generous support from the CUHK's Improvement on Competitiveness in Hiring New Faculties Funding Scheme and the CUHK's Direct Grant Project Number 4055191. 

\section*{Impact Statement}
Our paper presents an advancement in the field of machine learning and also has potential societal implications. By improving the understanding and capabilities of conditional distribution modeling, our framework can contribute to various real-world applications. For instance, it can aid in personalized treatment planning in healthcare by modeling conditional distributions of patient responses to different therapies. In finance, it can enhance risk assessment by predicting conditional distributions of asset returns under varying market conditions. By enabling more accurate and nuanced predictions in diverse domains, our research can positively impact decision-making processes, resource allocation, and, ultimately, the well-being of individuals and society.

\bibliography{biblio}
\bibliographystyle{icml2024}

\newpage
\appendix
\onecolumn

\section{Additional Experimental Details}
\label{sec:additional-details}
\subsection{Parameter Tuning}

We heuristically fix the learning rates $\alpha = \beta = 0.001$ as a reasonable value for almost every simple multilayer perception model. For other parameters, we conducted a grid search on minimizing the WD metric of the validation set. We found that the best combination for our parameters in the LDW-CPS dataset is KDE bandwidth of $h = 0.3$, $\varepsilon = 1.0$, $\lambda = 0.4$, $r_1 = 3.0$, $r_2 = 2.0$, $\gamma = 0.5$, $\delta = 0.7$. Regarding the ECM dataset, the best combination is $h = 0.2$, $\varepsilon = 1.0$, $\lambda = 0.4$, $r_1 = 3.0$, $r_2 = 2.0$, $\gamma = 0.5$, $\delta = 0.7$. We fix these parameter values for all our experiments in the main paper.

\subsection{Comparison between GENTLE and Baselines on Other Metrics}
In this section, we report two additional metrics, including Mean Squared Error (MSE) and $R^2$ score, to evaluate the performance of GENTLE and baselines. Tables~\ref{tab:addi_metrics_cps} and~\ref{tab:addi_metrics_ecm} represent the average results of algorithms across ten different seed numbers in terms of these two metrics. We observe that GENTLE outperforms baselines on both MSE and $R^2$ criteria. Although MGAN has comparable results to GENTLE in terms of MSE and $R^2$, the qualitative results shown in Figures~\ref{fig:qual1} and~\ref{fig:qual2} and the statistical metrics of WD and KS in Tables~\ref{tab:generalization} reveals its trend of collapsing to the mean of the ground truth distribution.

\begin{table}[h]
\centering
\caption{MSE and $R^2$ on generated $\hat{Y}(x)$ using different methods vs.~corresponding ground truth on the LDW-CPS dataset. The symbol $\pm$ represents the standard deviation.}
\fontsize{8}{12}\selectfont
\begin{tabular}{cccc}
\hline
Method & MSE ($\times 10^6$)($\downarrow$) & $R^2$($\uparrow$) \\ \hline
GENTLE & \textbf{3.70 $\pm$ 0.73} & \textbf{0.92 $\pm$ 0.05} \\
CWGAN & 108.43  $\pm$ 42.93 & -1.83  $\pm$ 0.13 \\
MGAN & 3.82  $\pm$ 0.71 & \textbf{0.92  $\pm$ 0.09} \\
WGAN-GP & 59.64  $\pm$ 9.65 & -0.59  $\pm$ 0.07 \\
CDSB & 8.54  $\pm$ 1.21 & 0.77  $\pm$ 0.08 \\ \hline 
\end{tabular}
    \label{tab:addi_metrics_cps}
\end{table}

\begin{table}[h]
\centering
\caption{MSE and $R^2$ on generated $\hat{Y}(x)$ using different methods vs.~corresponding ground truth on the ECM dataset. The symbol $\pm$ represents the standard deviation.}
\fontsize{8}{12}\selectfont
\begin{tabular}{cccc}
\hline
Method & MSE ($\downarrow$) & $R^2$($\uparrow$) \\ \hline
GENTLE & \textbf{0.62 $\pm$ 0.32} & \textbf{0.96 $\pm$ 0.05} \\
CWGAN & 24.30  $\pm$ 4.36 & 0.00  $\pm$ 0.13 \\
MGAN & 1.24  $\pm$ 0.22 & 0.91  $\pm$ 0.10 \\
WGAN-GP & 2.11  $\pm$ 0.33 & 0.86  $\pm$ 0.08 \\
CDSB & 2.01  $\pm$ 0.39 & 0.89  $\pm$ 0.04 \\ \hline 
\end{tabular}
    \label{tab:addi_metrics_ecm}
\end{table}

\subsection{Additional Experimental Results}

\begin{figure}[h]
  \centering
  \begin{subfigure}[b]{0.4\textwidth}
    \includegraphics[width=0.9\textwidth]{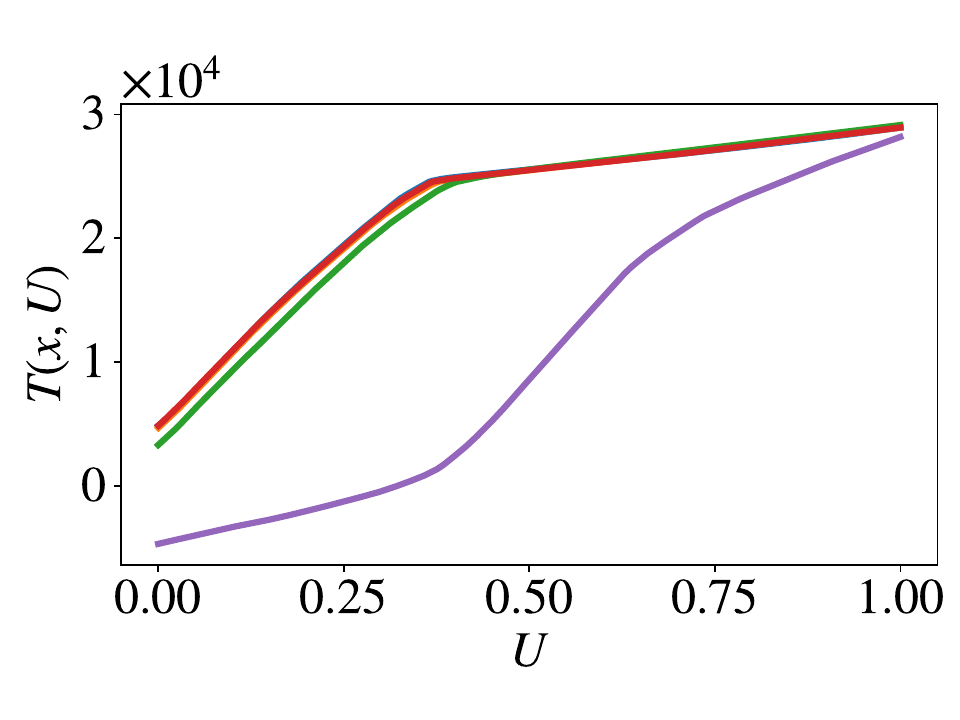}
    \caption{GENTLE trained on LDW-CPS}
    \label{fig:mono_cps}
  \end{subfigure}
  \begin{subfigure}[b]{0.4\textwidth}
    \includegraphics[width=0.9\textwidth]{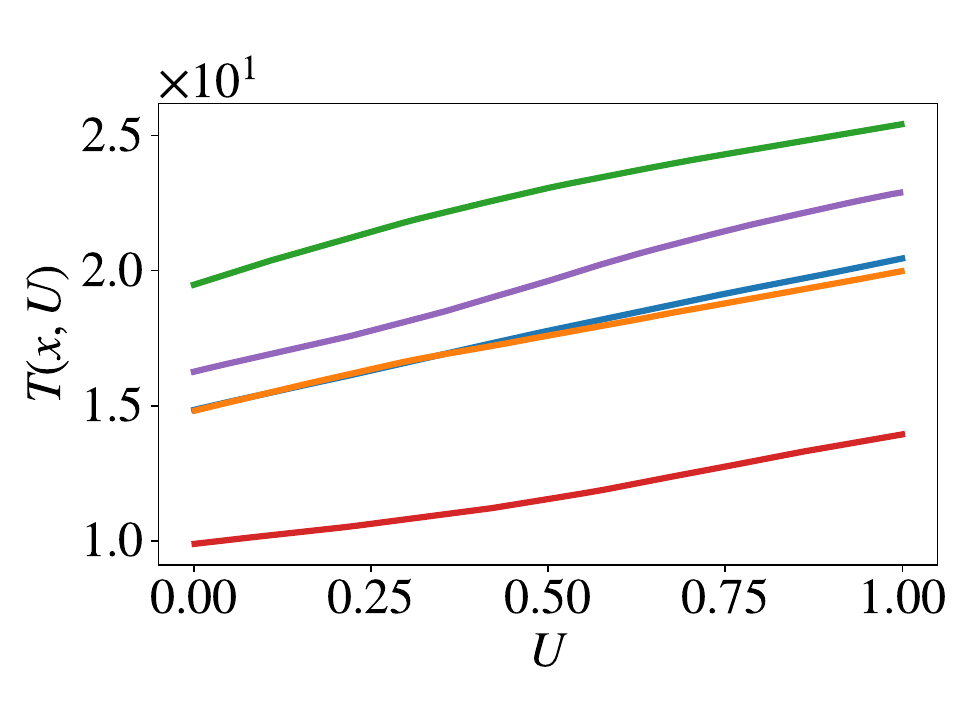}
    \label{fig:mono_cancer}
    \caption{GENTLE trained on ECM}
  \end{subfigure}
  \caption{Empirical evidence for the monotonically increasing property of the learned network $T_\theta(x, U)$ in the variable $U$.}
  \label{fig:mono}
\end{figure}

In Figure~\ref{fig:mono}, we plot the value of $T_\theta(x, U)$ for an increasing grid of $U \sim \cU(0, 1)$. We observe that the output of $T_\theta$ also increased with $U$, empirically confirming that $T_\theta$ is an increasing function, which conforms with the fact that $T_\theta(x, U)$ is an approximation of an increasing function of $U$.

\begin{figure*}[t]
     \centering
     \includegraphics[width=0.9\textwidth]{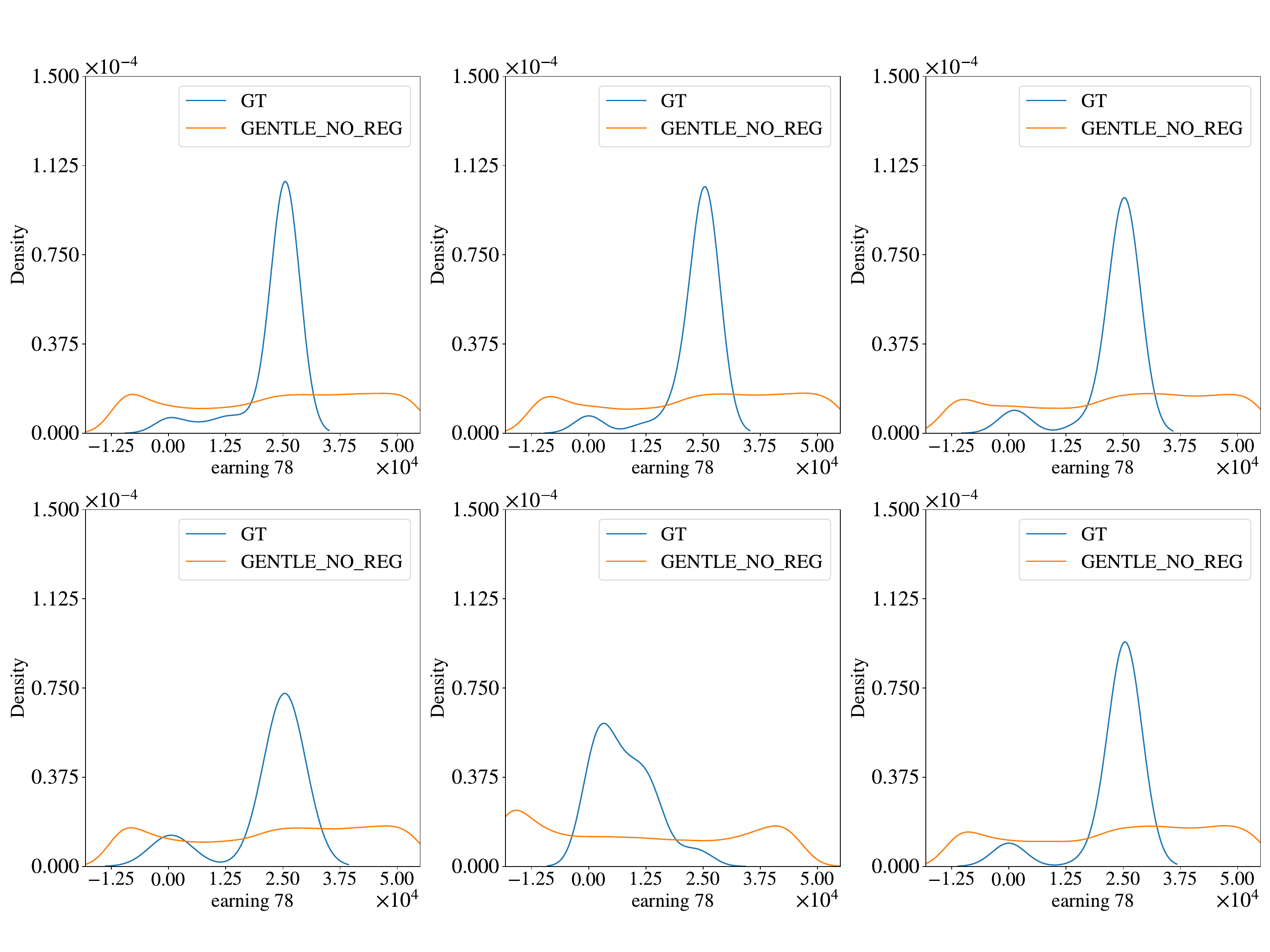}
     \caption{Qualitative result of the GENTLE version without regularization term on the LDW-CPS dataset. Each subfigure represents the results of a different covariate value in the test set. The learned distribution of this version tends to span over the range of the response term uniformly.}
     \label{fig:no_reg}
 \end{figure*}

 \begin{figure*}[t]
     \centering
     \vspace{-2mm}
     \includegraphics[width=0.85\textwidth]{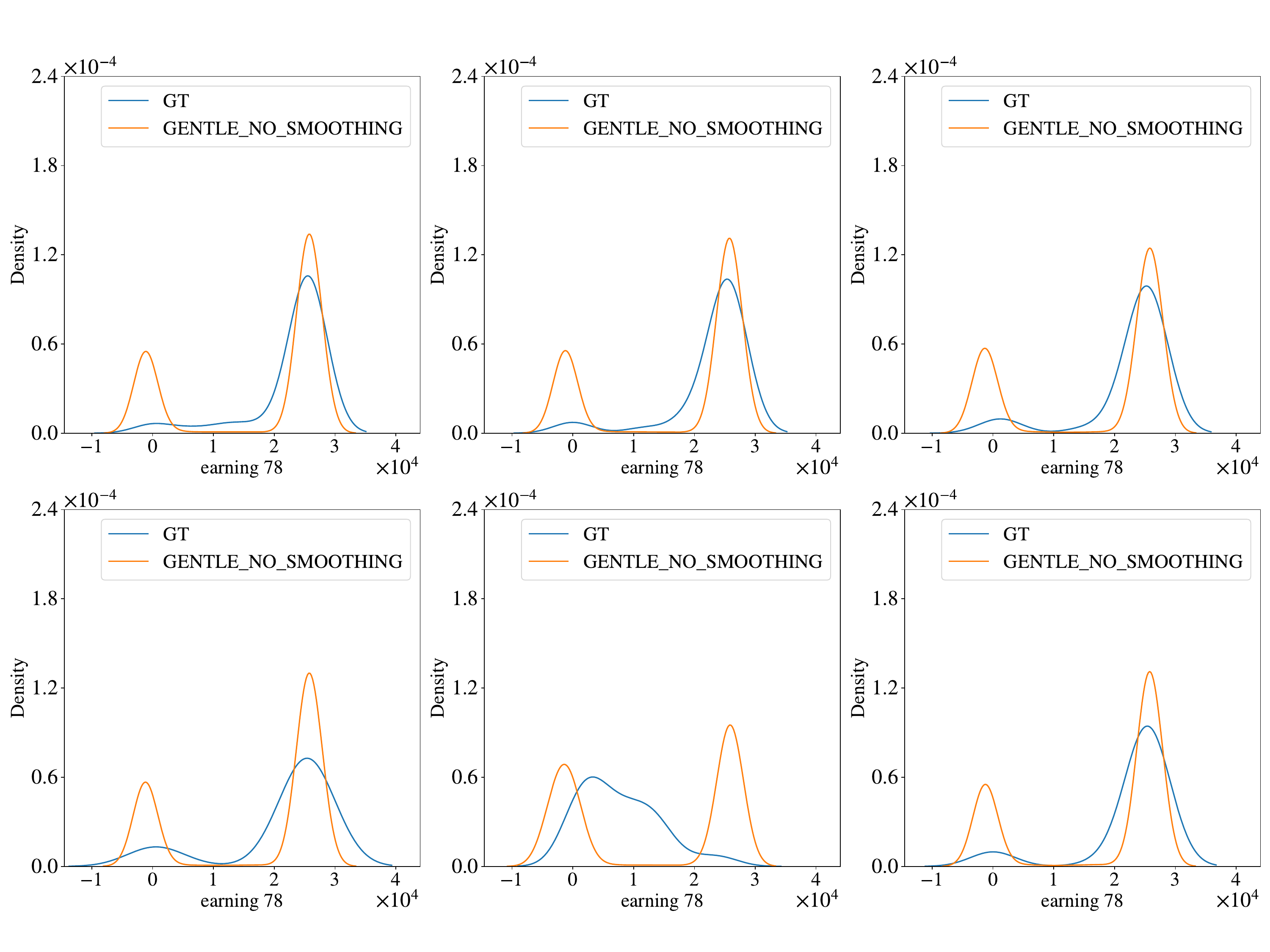}
      \vspace{-5mm}
     \caption{Qualitative result of the GENTLE version without smoothing term on the LDW-CPS dataset. Each subfigure represents the results of a different covariate value in the test set.}
     \label{fig:no_smoothing}
 \end{figure*}

  \begin{figure*}[t]
     \centering
     \vspace{-2mm}
     \includegraphics[width=0.85\textwidth]{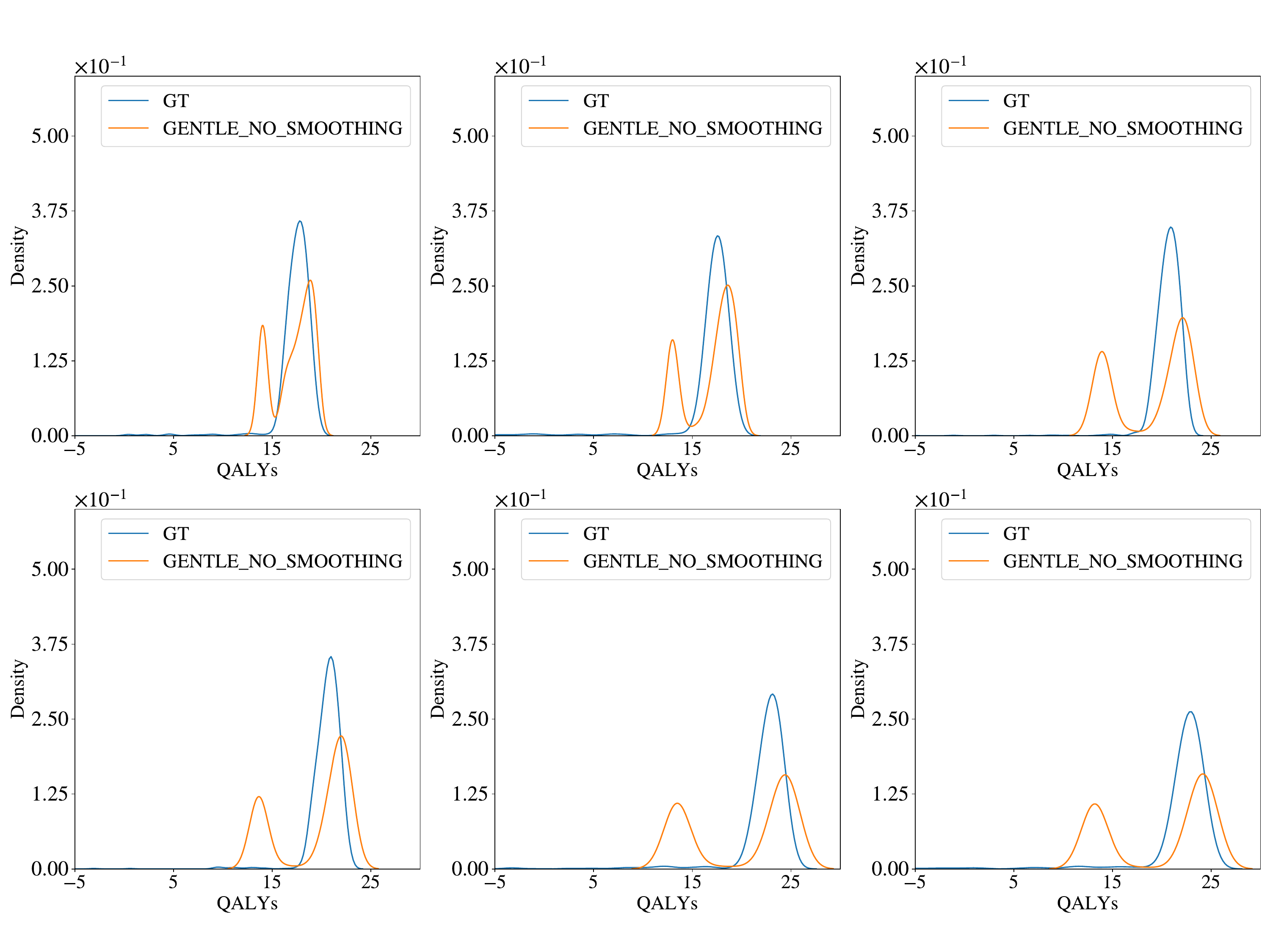}
     \vspace{-5mm}
     \caption{Qualitative result of the GENTLE version without smoothing term on the ECM dataset. Each subfigure represents the results of a different covariate value in the test set.}
     \label{fig:ecm_no_smoothing}
 \end{figure*}

\end{document}